\pdfoutput=1
\documentclass[11pt]{article}

\usepackage{acl}

\usepackage{times}
\usepackage{latexsym}
\usepackage{tabularx,booktabs}
\usepackage{tabu}
\usepackage{graphicx}
\usepackage{xcolor}
\usepackage{multirow}
\usepackage{subcaption}
\usepackage{colortbl}
\usepackage{float}

\usepackage[T1]{fontenc}
\usepackage[utf8]{inputenc}

\usepackage{microtype}
\title{Learning to Generate Prompts for Dialogue Generation through Reinforcement Learning}

\author{Hsuan Su$^*$,  Pohan Chi$^*$, Shih-Cheng Huang$^*$, Chung Ho Lam$^*$ \\ 
    {\bf Saurav Sahay$^+$}, {\bf Shang-Tse Chen$^{*\diamond}$}, {\bf Hung-yi Lee$^*$} \\
        {\bf National Taiwan University$^*$} \\
        {\bf Intel Labs$^+$} \\
        \textit {\{f09922053, r08942074, r09942093, r10921105, hungyilee\}@ntu.edu.tw$^*$} \\
        \textit{stchen@csie.ntu.edu.tw $^{\diamond}$ } \\
        \textit{saurav.sahay@intel.com$^+$ }
        }
\begin{document}
\maketitle
\begin{abstract}
% Much literature have shown that prompt-based learning is an efficient method to steer the output of a dialogue model.  Gradient-based learning is one of the well-known methods that could access models’ parameters to perturb the prompts. However, language models nowadays are growing larger, making them too ponderous to backpropagate. Some of them are not even available to the public.
% In this work, we first explored the combination of prompting and reinforcement learning (RL) to steer models’ generation without accessing any of the models' parameters.
% Second, to reduce the training effort and enhance the generalizability to the unseen task, we applied multi-task learning to make the model learn to generalize to new tasks. 
% The experiment results show that our proposed method successfully controlled several state-of-the-art(SOTA) dialogue models without taking a look at their parameters. Furthermore, the model also demonstrated the strong ability to quickly adapt to an unseen task in fewer steps compared to the baseline model.
Much literature has shown that prompt-based learning is an efficient method to make use of the large pre-trained language model. Recent works also exhibit the possibility of steering a chatbot's output by plugging in an appropriate prompt. Gradient-based methods are often used to perturb the prompts. However, some language models are not even available to the public. In this work, we first explored the combination of prompting and reinforcement learning (RL) to steer models' generation without accessing any of the models' parameters.
Second, to reduce the training effort and enhance the generalizability to the unseen task, we apply multi-task learning to make the model learn to generalize to new tasks better. 
The experiment results show that our proposed method can successfully control several state-of-the-art (SOTA) dialogue models without accessing their parameters. Furthermore, the model demonstrates the strong ability to quickly adapt to an unseen task in fewer steps than the baseline model.
\end{abstract}

\section{Introduction}
Prompt-based learning \cite{liu2021pretrain} is an efficient method to utilize extremely large pre-trained language models. Unlike traditional supervised learning, prompt-based learning does not predict the label directly. To use the pre-trained language model to do prediction tasks, a prompt template describing the downstream task would be constructed. Then the original input would be filled into the input slot in the prompt template. Such method was proved effective in previous works \cite{mccann2018natural, radford2019language, schick2020exploiting}. 

\cite{liu2021gpt, liu2021p} studied the automatic generation of prompts. The downstream tasks were solved by adding automatically generated prompts to the inputs. Some \cite{li2021prefix} had also explored ways to adapt prompting in NLG tasks. The generation was controlled by modifying the prompts added as a prefix to the input. The parameters used to generate the prompts were updated with gradient-based methods. 
In \cite{sheng-etal-2020-towards}, they aimed to generate the prompts to  induce negative biases for one demographic and positive biases for another demographic and equalized biases between demographics with universal triggers~\cite{Wallace2019Triggers}.

However, the parameters of some language models were not publicly available. For example, the powerful GPT-3 \cite{brown2020language} model was not open sourced, making it impossible for researchers to access its parameters or apply gradient-based methods.
It is important to develop a technique that influence model's output without accessing its parameters. \cite{su-etal-2021-put} proposed a method to train a chatbot that can influence interlocutor's response with policy gradient~\cite{silver2014deterministic}. Their proposed method can prolong the interlocutor's response, induce the interlocutor to respond with emotion and specific words.

\begin{figure}
    \centering
    \includegraphics[width = \linewidth]{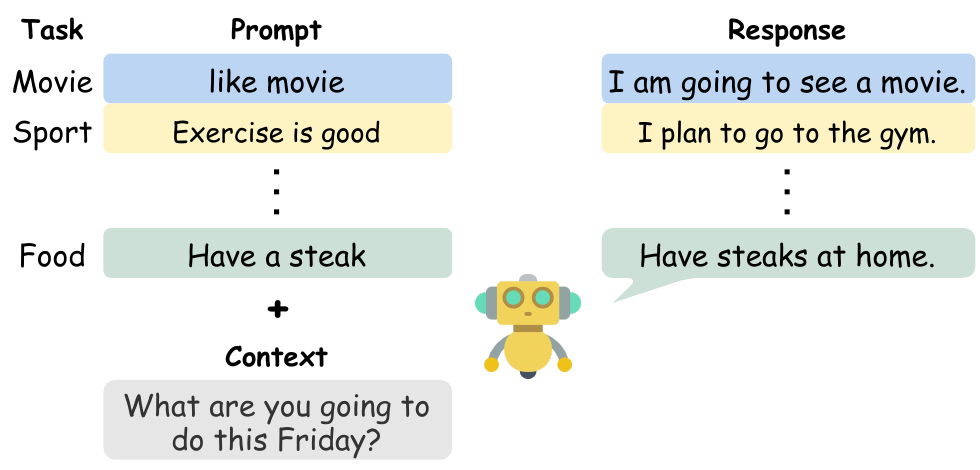}
    \caption{Steering model's output with prompts}
    \label{fig:framework_small}
\end{figure}

With these in my mind, we proposed a new framework that combined RL and prompt-based learning to steer the output of any off-the-shelf model without accessing its parameters. As figure \ref{fig:framework_small} showed, we could generate different prompts for each task and concatenate it with the original input sentence. Then the modified sentence was taken as the input of the conversational model to influence the response. More details would be explained in section \ref{ssec:framework}

Our contributions are:
\begin{itemize}
  \item We proposed a novel framework that used reinforcement learning to steer off-the-shelf conversational model to reply with emotion and specific topic without accessing any of its parameters.
  \item Our proposed framework can not only enhance the generalizability toward unseen tasks, it also can decrease the training steps on the unseen tasks efficiently.
  \item We further conduct the human trails and applied our framework on the state of the art GPT3 and successfully controlled it generated response. 
\end{itemize}

\section{Related Work}
\subsection{Controllable Generation}
Controllable language generation becomes gradually popular in recent years. \cite{keskarCTRL2019} proposed a 1.6 billion-parameter conditional transformer language model conditioning on control codes that specify domain, subdomain, entities, relationships between entities, dates, and task-specific behavior. \cite{Dathathri2020Plug} used the gradient-based method that allows a user to flexibly plug in one or more tiny attribute models representing the desired steering objective into a large, unconditional language model (LM). The LM was influenced by the gradient from the plugged classifier.
Also, many of their works also want to influence model's output without accessing its parameters but only take the model's output probabilities.
\cite{KrauseGeDi2020, liu-etal-2021-dexperts} both influenced the models' output with multiplying an external attribute probability on the original LM probabilties using Bayes Rule.
\cite{ziegler2019finetuning} fine-tuned the GPT-2 model's on human preference to control GPT-2's output with reinforcement learning.
\subsection{Few-shot Learning for Dialogue Generation}
Much literature combined MAML\citep{pmlr-v70-finn17a} with transformer\citep{NIPS2017_3f5ee243} models to enhance the fast adaptation ability of chatbot. \citep{madotto-etal-2019-personalizing} proposed a method that extended MAML to personalized dialogue learning; the chatbot could learn to quickly adapt to new personas by leveraging only a few dialogues.
\cite{zhu-etal-2019-multi} first proposed propose a novel multi-task learning framework for Dialogue State Tracking (DST) to steer mode's output in various dialogue acts.
\citep{mrksic-etal-2017-neural} applied the multi-task learning algorithm to DST and the obtained meta-learning model is used for new domain adaptation.
\citep{gao-etal-2020-dialogue-generation} combined structure learning into chatbot to enhance the effectiveness of meta-learning by promoting knowledge customization among different sentence functions. 
These previous works show the power of the MAML and multi-task learning that when the chatbot model encounters an unseen task, it still can quickly adapt to the new tasks.

\section{Method}

\begin{figure*}
    \centering
    \includegraphics[width = \linewidth]{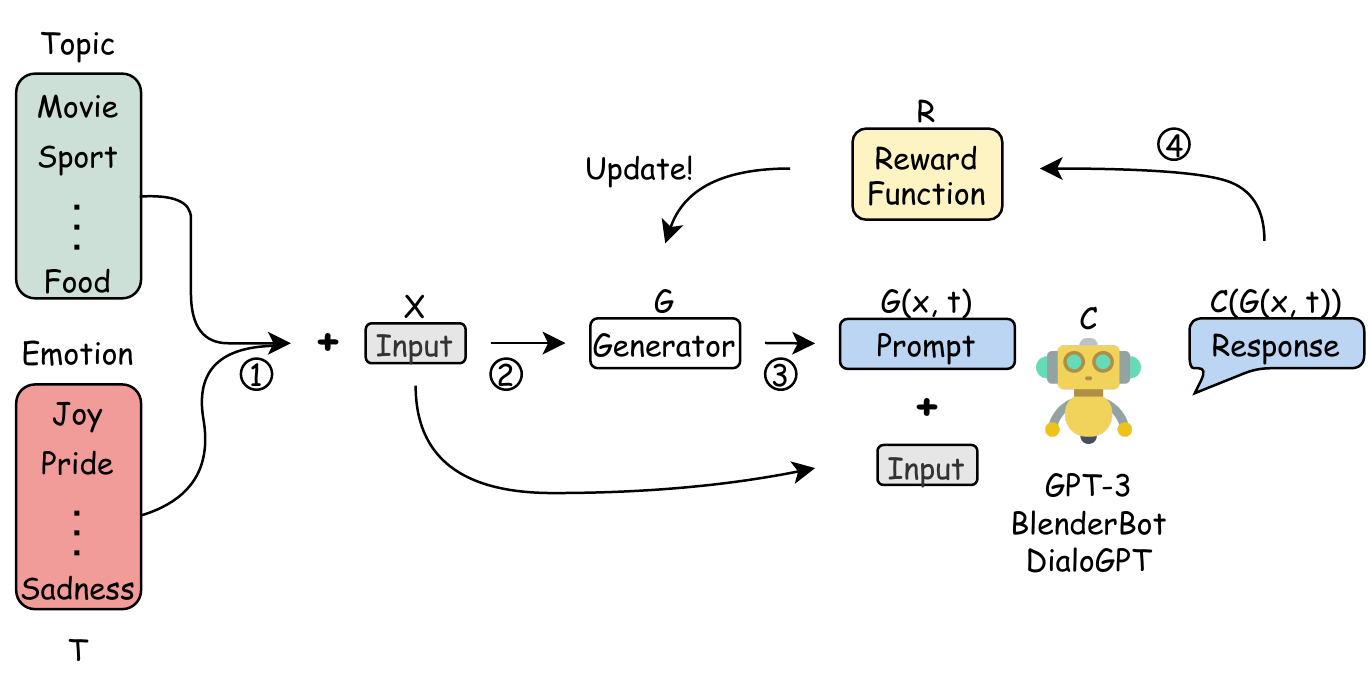}
    \caption{The proposed framework. (1) The controllable factor is concatenated to the input. (2) The generator takes the modified sentence as the input. (3) The generator generates a prompt according to the input. (4) The conversational model generates the response from the prompt and the input.}
    \label{fig:framework}
\end{figure*}

\subsection{Framework} \label{ssec:framework}
The proposed framework is shown in Figure \ref{fig:framework}.  It consisted of two models, a prompt generator and an conversational model. The prompt generator would take the original sentence and a controllable factor as its input and generate a prompt related to the controllable factor. The prompt was concatenated with the original sentence to form the input sentence of the conversational model. The generator aimed to produce the prompts that would steer the conversational model’s response, making the output correlated to the controllable factor. 

\subsection{Conversational Model}
The conversational model is denoted as $C$, which can be an arbitrary chatbot that we may or may not have access to its parameters. It means that $C$ can be any off-the-shelf model with fixed parameters or even just an API. It is because we train the prompt generator with RL, and there is no need to modify or backpropagate through the parameters of $C$.

\subsection{Prompt Generator} \label{ssec:prompt_generator}
The prompt generator $G$ was initialized with the DialoGPT \cite{zhang-etal-2020-dialogpt} model. 
We constructed the input of $G$ by concatenating the controllable factor and the original input sentence, denoted as follows:
\begin{equation}
s(c,x) = [<c>;x],
\end{equation}
where $c$ denotes the controllable factor, and $x$ denotes the original input sentence. When the controllable factor is the emotion, some of the possible choices of $c$ are \textit{joy}, \textit{sadness}, etc. When the controllable factor is the topic, $c$ can be \textit{sports}, \textit{musicalinstruments}, and so on. Then the generator $G$ takes $s$ as the input and generates the prompt $p = G(s)$. Finally, $p$ would be concatenated with the original input sentence $x$ to form the input of the conversation $C$. The final response is $C([p;x])$ generated by the conversational model $C$. 
We defined the reward function $R$ for $G$ based on $C([p;x])$. $G$ would learn to maximize $R$ by generating appropriate prompts to lead the output of $C$. The exact definition of the reward function $R$ depends on the type of the controllable factor. We applied RL to update the parameters of the model. Detailed explanation is described in section \ref{ssec:rl}. 

\subsection{Reinforcement Learning} \label{ssec:rl}
We apply the Proximal Policy Optimization (PPO) \cite{schulman2017proximal} as our RL training algorithm. The definition of the reward function $R$ depends on the the controllable factor. We have designed different reward functions respectively. 
% The reward $R$ was composed of two parts, the guiding reward $R_g$ and the coherence reward $R_c$. $R_c$ was defined as \textcolor{red}{(formula?)}. It was meant to preserve the fluency of the prompt. 

\subsubsection*{Emotion}
We aim to generate prompts that can steer the conversational model to respond with a specific emotion. The emotion taxonomy follows the setting in \cite{rashkin-etal-2019-towards}. There are in total 27 emotion categories, including positive and negative emotions. To identify the emotion embodied in the utterance, we build a BERT-based emotion classifier $B$ to determine the emotion category that the response belongs to. The emotion reward $R_E$ is defined as the following:
\begin{equation}
    R_E = B(y|C([p;x]))
\end{equation}
The emotion reward $R_E$ is the probability predicted by the classifier $B$. $I([p;x])$ is the response generated by the conversation, and $y$ is the given emotion.
\subsubsection*{Topic}
We attempt to induce the conversational model to say specific words relevant to a certain topic. For example, if the given topic $w$ is $sports$, the corresponding words will be $athlete$, $baseball$, $coach$, etc. We call the collection of the specific words a "word group". The topics and the corresponding word groups are collected from the EnchantedLearning website\footnote{\url{https://www.enchantedlearning.com/wordlist/}}. We crawl various topics from the vocabulary word lists, but some topics are filtered out due to their lack of semantic relationship. To encourage the conversation to respond with specific words, we count the frequency of the specific words that appear in the response sentence and use it as the reward function $R_W$. We expect that the conversation will learn to respond with sentences containing words relevant to the controllable factor.

To summarize, if the controllable factor is emotion, we let $R = R_E$; for the cases that the controllable factor is topic, $R = R_W$.

\subsection{Multi-task Learning}
We find that the reward converges very slowly with pure RL training. To make the model learn to adapt to new tasks faster, we apply multi-task learning to the training procedure. During training, for each controllable factor we randomly choose several tasks, and optimize the reward of these tasks together. By jointly training with multiple tasks, the hierarchical relationship can also be exploited implicitly. We anticipate that the model will not only learn to solve a single task but also obtain better generalizability.

\section{Experimental Setup}

\subsection{Dataset}
\subsubsection*{EmpatheticDialogues}
EmpatheticDialogues \cite{rashkin-etal-2019-towards} is a novel dataset containing around 25K conversations. Each conversation is grounded in a specific situation, where the speaker was feeling a given emotion, with a listener responding. We trained our proposed framework on this dataset.
\subsubsection*{GoEmotions}
GoEmotions is a dataset that contains 58k English Reddit comments created by \cite{demszky2020goemotions}. It is labeled for 27 emotion categories or Neutral, each comment contains one or multiple labels. These categories include a large number of positive, negative, and ambiguous emotions. The diversity of the emotion categories makes it suitable for downstream conversation understanding tasks. Additionally, it is by far the largest human annotated emotion classification dataset. \cite{demszky2020goemotions} also showed reliable dissociation among all 27 emotion categories, indicating the suitability for building an emotion classification model on it. We built a BERT-based emotion classifier trained with GoEmotions following the github repo \footnote{\url{https://github.com/google-research/google-research/tree/master/goemotions}} they released to predict the emotion reward $R_e$ for the emotion controllable factor.

\subsection{Model Settings}
\subsubsection*{Training Details}
We applied the PPO as our RL optimization algorithm. We also combined multi-task learning in our training process to improve generalization. For the topic, there were 95 tasks in the training set and 21 tasks in the testing set. Each topic was considered a single task. For the emotion, there were 21 tasks in the training set and 6 tasks in the testing set. Similarly, each emotion was considered a single task. In the multi-task training, we randomly sampled 8 tasks for each epoch and optimized the rewards simultaneously. After the training was finished, we tested the model on the unseen tasks under the few-shot setting.
\subsubsection*{Conversational Model}
We tested the proposed framework on three off-the-shelf conversational models.
\begin{itemize}
  \item The \textbf{DialoGPT} was the same model mentioned in section \ref{ssec:prompt_generator}. Here it acted as a conversational model with its weights fixed.
  To implement the prompt generator, we applied the DialoGPT model, which fine-tuned the GPT-2 model on 147M multi-turn dialogues from Reddit discussion threads. The GPT-2 model is a transformer-based model with 36 layers, 20 attention heads in each layer, 345M parameters, and an embedding size is 1024. This model is trained on the WebText dataset and 50,257 tokens with invertible byte pair encoding to preserve capitalization and punctuation. 
  \item The \textbf{Blenderbot} \cite{roller-etal-2021-recipes} is an open-domain chatbot released by Facebook AI. It was trained and evaluated on the Blended Skill Talk (BST) \cite{smith2020can} task and was able to show engagement and empathy in the conversation.
  \item The \textbf{GPT-3} \cite{brown2020language} is a large language model with 175B parameters. It can achieve strong performance on a wide variety of natural language tasks without any gradient updates or fine-tuning, but the model parameters are not available to the public. Luckily, OpenAI's API provided access to GPT-3, allowing us to evaluate our framework on it. Due to the quota limit and the cost, we only tested the model on two tasks, office for topic and pride for emotion. The model we used was Ada, the fastest GPT-3 model released by OpenAI. 
\end{itemize}

\subsection{Baseline}
We compared the proposed framework with the following baselines:
\subsubsection*{No Prompt}
The context $x$ was directly taken by the conversational model $C$ to generate the response $C(x)$.
\subsubsection*{DialoGPT}
We also trained a model that initted with the pre-trained DialoGPT directly without multi-task learning. This method is denoted as DGPT in the following experiments.

\subsubsection*{Independent Prompt}
In order to find out whether the input sentence was necessary, we tried generating prompts without relying on the context $x$. In other word, the prompt $p = G(s(c)) = G(<c>)$ was determined only by the controllable factor $c$.
\subsubsection*{Human Prompt}
We manually designed two prompt templates to compare with the model-generated prompts. The handcrafted prompts were \textit{"There is \{c\} in the following response: "} and \textit{"Making the following response full of \{c\}: "}, denoted as \textit{human-0} and \textit{human-1} respectively. The \textit{\{c\}} in the template was the placeholder for the controllable factor.

\subsection{Evaluation Metrics}
\subsubsection*{Perplexity}
Here we employed the pretrained GPT-2 language model $GPT$ to judge if the output sentence $C(x)$ was an acceptable sentence. The computation of Perplexity \citep{chen1998evaluation} is shown below.
\begin{equation}
  PPL = \prod_{i=1}^{T} \frac{1}{(GPT(C(x, D)_i|x))^{1/T}}
\end{equation}

\subsubsection*{Coherence}
We employed the DialogRPT \citep{gao2020dialogrpt} to calculate the coherence between conversation model's output and the input context. DialogRPT \citep{gao2020dialogrpt} is a GPT2-based ranker that finetuned on 133M human feedback data. With the contrastive learning approach that DialogRPT used. The ranker has better understanding on how relevant the response is for the given context. In our evaluation, we take the the probability that output by DialogRPT coherence model (\textit{human\_vs\_rand}) as our coherence metric.
\subsubsection*{Self-BLEU}
BLEU score \cite{papineni2002bleu} is a commonly used metric for automatically evaluating machine translation. However, the Self-BLEU \cite{zhu2018texygen} score here was applied to measure the diversity of chatbot responses. Regarding one sentence as the prediction and the others as the reference, we can calculate BLEU score for every sentence, and the average is the Self-BLEU score. A lower Self-BLEU score implies more diversity of the chatbot responses.

\begin{figure*}[h]
    \begin{subfigure}{0.5\linewidth}
        \includegraphics[width=\linewidth]{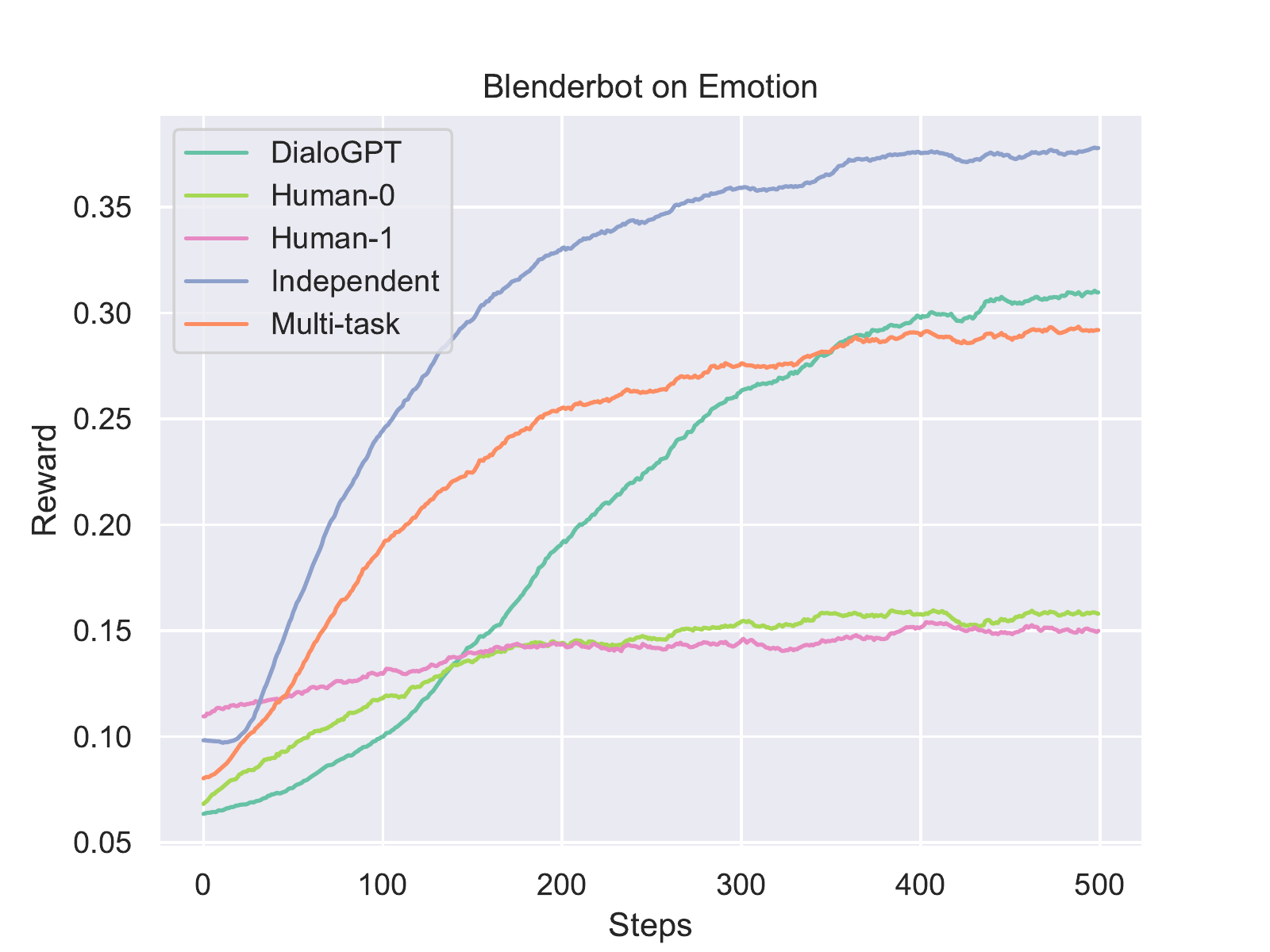} 
        \caption{Blenderbot on Emotion}
        \label{fig:subim21}
    \end{subfigure}
    \hfill
    \begin{subfigure}{0.5\linewidth}
        \includegraphics[width=\linewidth]{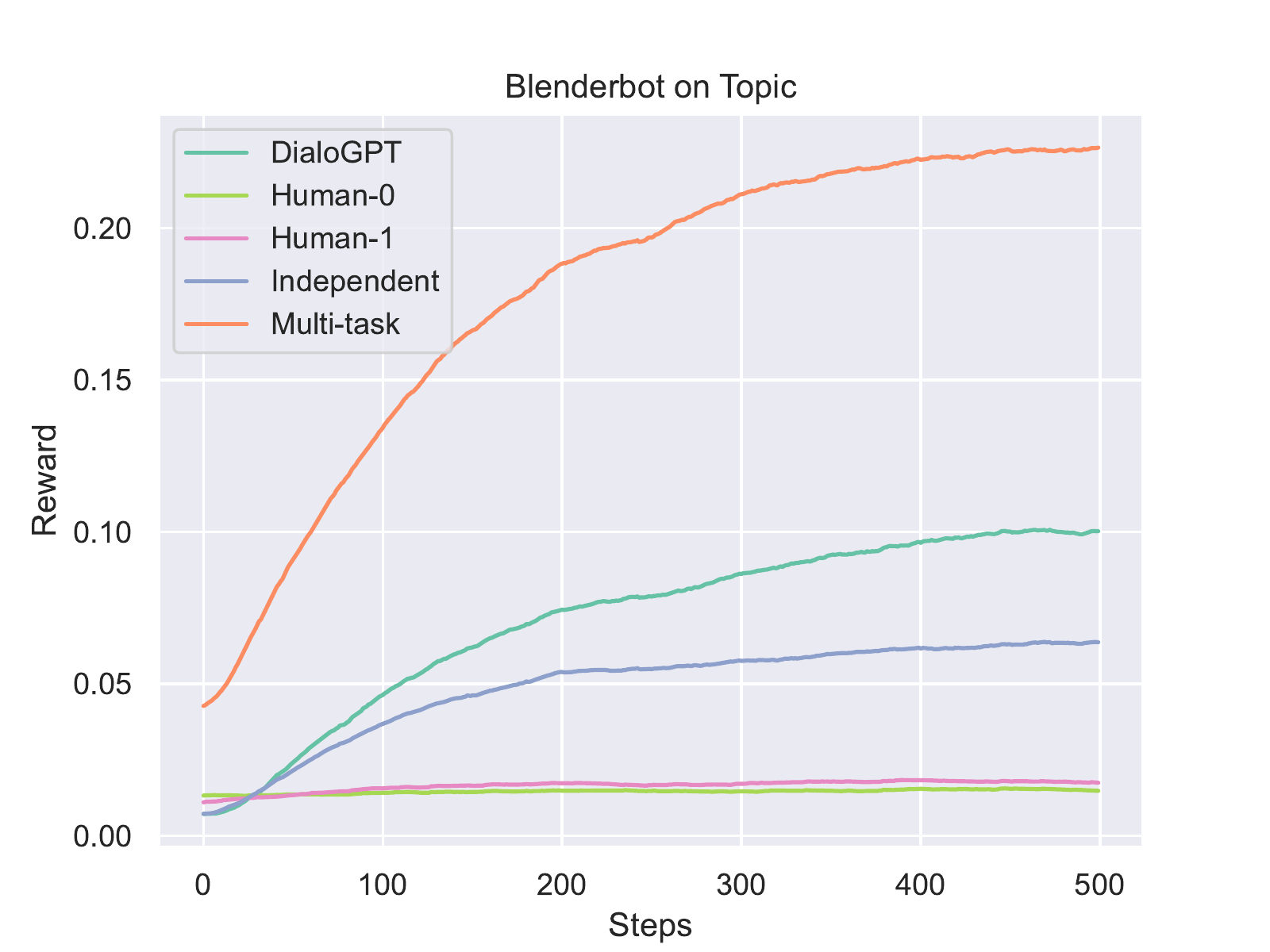}
        \caption{Blenderbot on Topic}
        \label{fig:subim22}
    \end{subfigure}
    \begin{subfigure}{0.5\linewidth}
        \includegraphics[width=\linewidth]{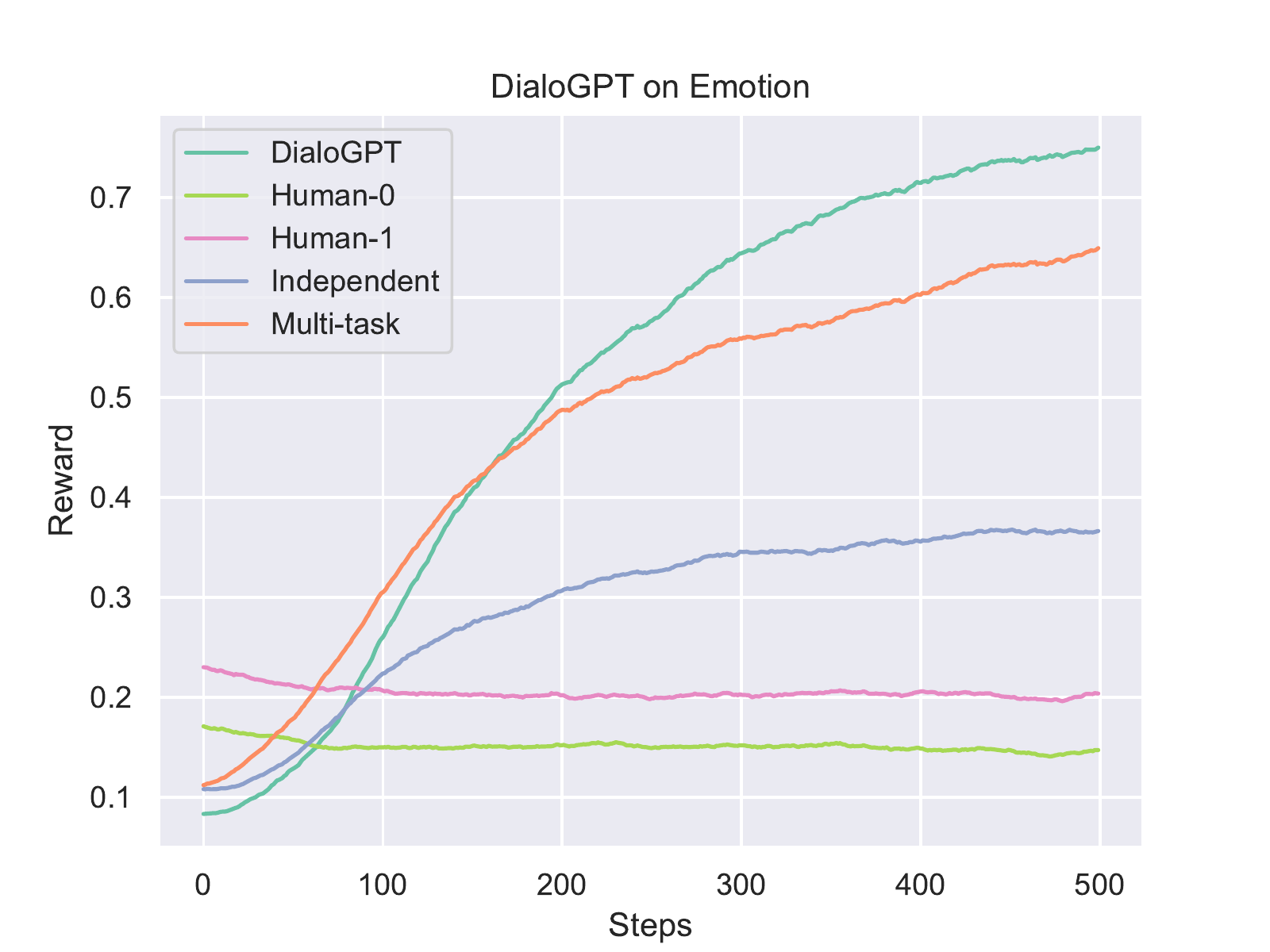} 
        \caption{DialoGPT on Emotion}
        \label{fig:subim23}
    \end{subfigure}
    \hfill
    \begin{subfigure}{0.5\linewidth}
        \includegraphics[width=\linewidth]{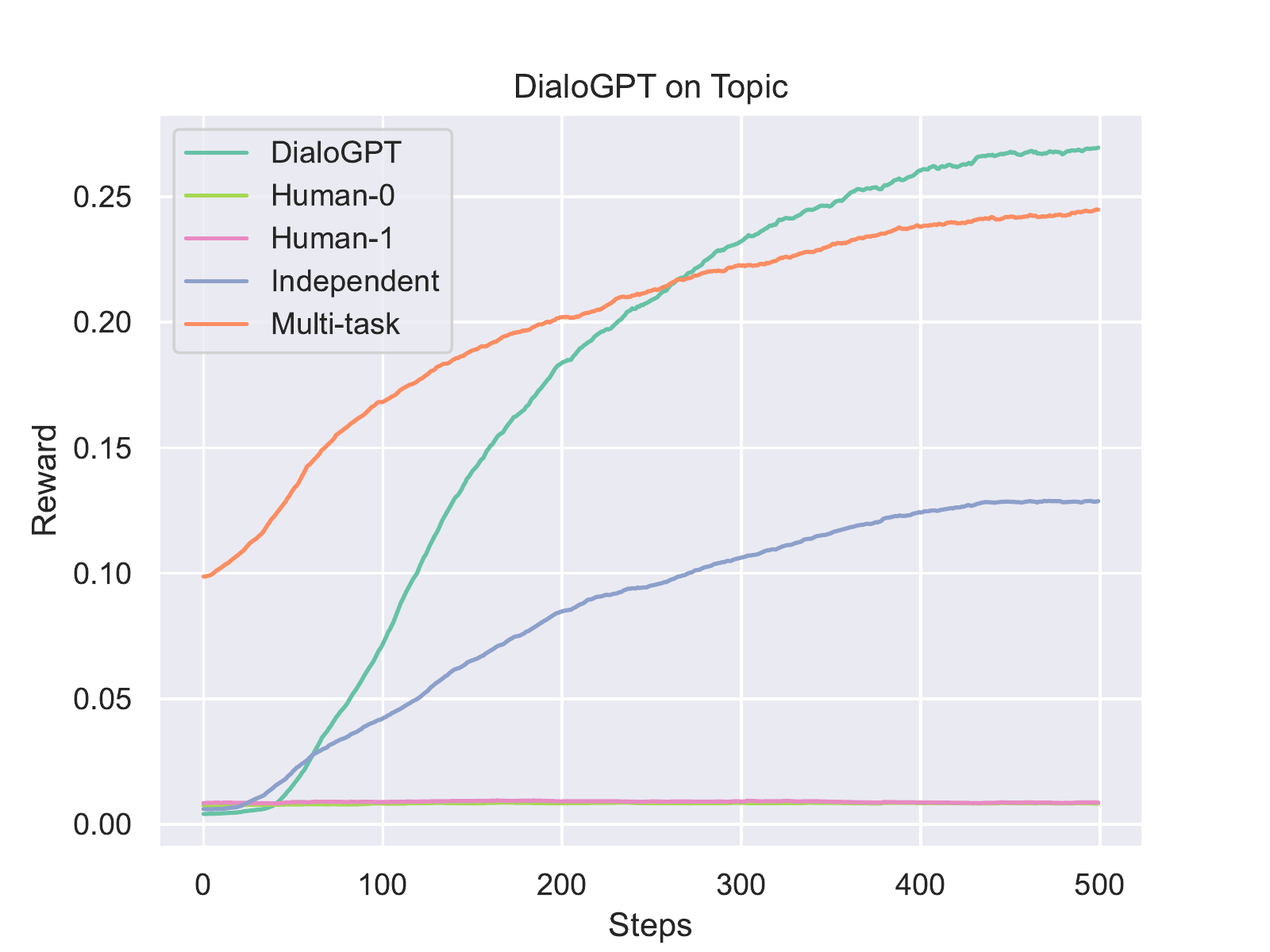}
        \caption{DialoGPT on Topic}
        \label{fig:subim24}
    \end{subfigure}
    \caption{Experiments on different controllable factors and conversational models. DialoGPT stands for the pre-trained only model; human-0 and human-1 represent the handcrafted prompts; independent refers to the prompts generated without input; multi-task is the proposed framework.}
    \label{fig:image2}
\end{figure*}

\begin{table*}[htbp]\small
        \centering
        \begin{tabular}{ccccccccccc}
        \hline
         &  &  &  & \multicolumn{2}{c}{\textbf{Emotion}} &  &  & \multicolumn{2}{c}{\textbf{Topic}} &  \\ \hline
         & \textbf{Method} & \textbf{Model} & \textbf{Reward} & \textbf{PPL} & \textbf{Coh.} & \textbf{SB-2} & \textbf{Reward} & \textbf{PPL} & \textbf{Coh.} & \textbf{SB-2}\\ \hline
         \multirow{9}{*}{\textbf{DialoGPT}}  & \multirow{2}{*}{0 shot} & DGPT & 0.092 & 19.048 & 0.665 & 0.869 & 0.005 & 19 & 0.655 & 0.867 \\
         &  & Multi & \textbf{0.133} & 18.785 & 0.657 & 0.896 & \textbf{0.092} & 20.149 & 0.412 & 0.923 \\
         & \multirow{2}{*}{10 shot} & DGPT & 0.107 & 19.011 & 0.651 & 0.88 & 0.007 & 18.482 & 0.636 & 0.879\\
         &  & Multi & \textbf{0.183} & 19.945 & 0.654 & 0.902 & \textbf{0.156} & 19.327 & 0.393 & 0.938\\
         & \multirow{2}{*}{500 shot} & DGPT & \textbf{0.747} & 8.684 & 0.39 & 0.942 & \textbf{0.263} & 14.442 & 0.448 & 0.933\\
         &  & Multi & 0.653 & 25.218 & 0.441 & 0.941 & 0.249 & 17.986 & 0.37 & 0.949\\ \cline{2-11}
         & Human-0 & DGPT & 0.156 & 16.068 & 0.635 & 0.904 & 0.008 & 15.17 & 0.567 & 0.911 \\
         & Human-1 & DGPT & 0.201 & 16.228 & 0.646 & 0.909 & 0.009 & 16.102 & 0.604 & 0.908 \\ \cline{2-11}
         & no prompt & - & 0.089 & 20.371 & 0.749 & 0.887 & 0.004 & 20.371 & 0.749 & 0.887  \\ \hline
        \multirow{9}{*}{\textbf{Blenderbot}}  & \multirow{2}{*}{0 shot} & DGPT & 0.078 & 16.64 & 0.823 & 0.918 & 0.008 & 16.756 & 0.823 & 0.916\\
         &  & Multi & \textbf{0.098} & 16.61 & 0.824 & 0.922  & \textbf{0.061} & 19.954 & 0.647 & 0.928\\
         & \multirow{2}{*}{10 shot} & DGPT & 0.082 & 16.316 & 0.821 & 0.922 & 0.019 & 17.321 &0.808 & 0.92\\
         &  & Multi & \textbf{0.147} & 16.452 & 0.823 & 0.926 & \textbf{0.112} & 21.992 &0.57 & 0.943\\
         & \multirow{2}{*}{500 shot} & DGPT & \textbf{0.312} & 15.322 & 0.752 & 0.944 & 0.1 & 20.059 & 0.671 & 0.942\\
         &  & Multi & 0.288 & 16.78 & 0.809 & 0.935 & \textbf{0.227} & 21.977 & 0.482 & 0.954\\ \cline{2-11} 
         & Human-0 & DGPT & 0.158 & 17.531 & 0.851 & 0.93 & 0.015 & 17.776 & 0.852 & 0.928\\
         & Human-1 & DGPT & 0.149 & 18.179 & 0.864 & 0.93 & 0.017 & 19.129 & 0.866 & 0.925\\ \cline{2-11}
         & no prompt & - & 0.069 & 18.141 & 0.877 & 0.925 & 0.007 & 18.141 & 0.877 & 0.925 \\ \hline
        \end{tabular}

    \caption{\label{tab:table1}The result of few-shot testing on DialoGPT and Blenderbot. The column \textbf{Model} means which model the prompt generator was initialized from.}
\end{table*}

\subsection{Human Evaluation}
Here we used DialoGPT as the conversational model to generate the results. We evaluated our results by hiring US workers on Amazon Mechanical Turk (MTurk). There were totally 28 testing tasks (22 for topic, 6 for emotion), and 5 testing results were sampled from each task. Each result was evaluated by 3 workers, so we collected 330 human evaluation results for topic and 90 results for emotion. Every worker was given a context sentence (the original input $x$) and two corresponding replies generated by the proposed framework and the baseline respectively. The workers were asked to compare one reply against the other in three aspects: fluency, coherence and topic/emotion score. Fluency means how natural the response is. Coherence represents whether the response is related to the context sentence. Topic/emotion score refers to how much the reply is relevant to the given controllable factor.

\section{Discussion and Analysis}
\subsection{Multi-task Learning}
We apply multi-task learning in order to make the model adapt to the unseen tasks faster. The experiment result is shown in Figure \ref{fig:image2}. The figures show that multi-task learning successfully steers both Blenderbot and DialoGPT to generate responses with higher rewards in few steps. Though multi-task learning did not achieve the highest reward in the end every time, what we focus on is the few-shot performance. In Figure \ref{fig:subim22} and \ref{fig:subim24}, multi-task learning show extraordinary potential in few-shot adaptation on unseen tasks. We speculate that multi-task learning can leverage useful information contained in multiple related tasks, such as the hierarchical relationship between the controllable factor and the corresponding tasks. For example, the relationship (\textit{food}, \textit{steak}) can be implied from another task (\textit{sports}, \textit{baseball}). This makes the model perform quite well even with only few training samples or even zero-shot testing.

\subsection{Human and Independent Prompt}
In Figure \ref{fig:subim22} and \ref{fig:subim24}, both of the human prompts hardly contribute to the reward, and the performance of the independent prompt is worse than the multi-task one. Nevertheless, we observe that the independent prompt outperforms the others in Figure \ref{fig:subim21}. 
\subsection{Influence of Few-shot Samples}
We investigate how the number of few-shot samples affect the performance. We compare the results of different shots in Table \ref{tab:table1}. The proposed model outperforms the baseline model in 0 shot and 10 shot scenarios for both controllable factors and conversational models. However, we also observe that the baseline model often surpasses the proposed model after training with 500 shots. From the observation, we can infer that multi-task learning do not always lead to better performance when the data is enough, which is also an interesting issue that we will study in the future.
% It implied that better ability to generalize and fast adapt didn't always lead to better performance when data was enough.

\begin{figure*}[h]
    \begin{subfigure}{0.5\linewidth}
        \includegraphics[width=\linewidth]{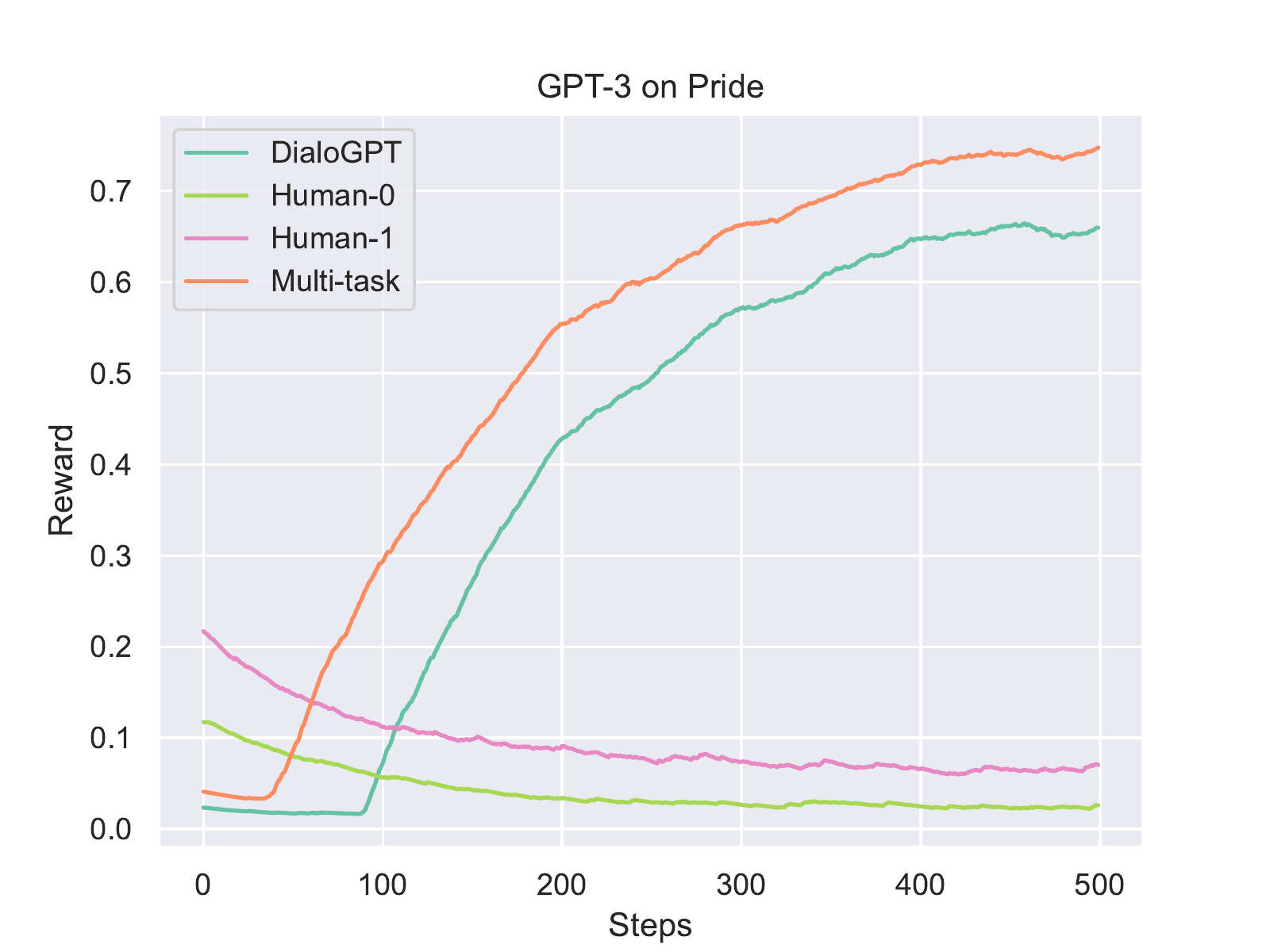} 
        \caption{GPT-3 on Pride}
        \label{fig:subim41}
    \end{subfigure}
    \hfill
    \begin{subfigure}{0.5\linewidth}
        \includegraphics[width=\linewidth]{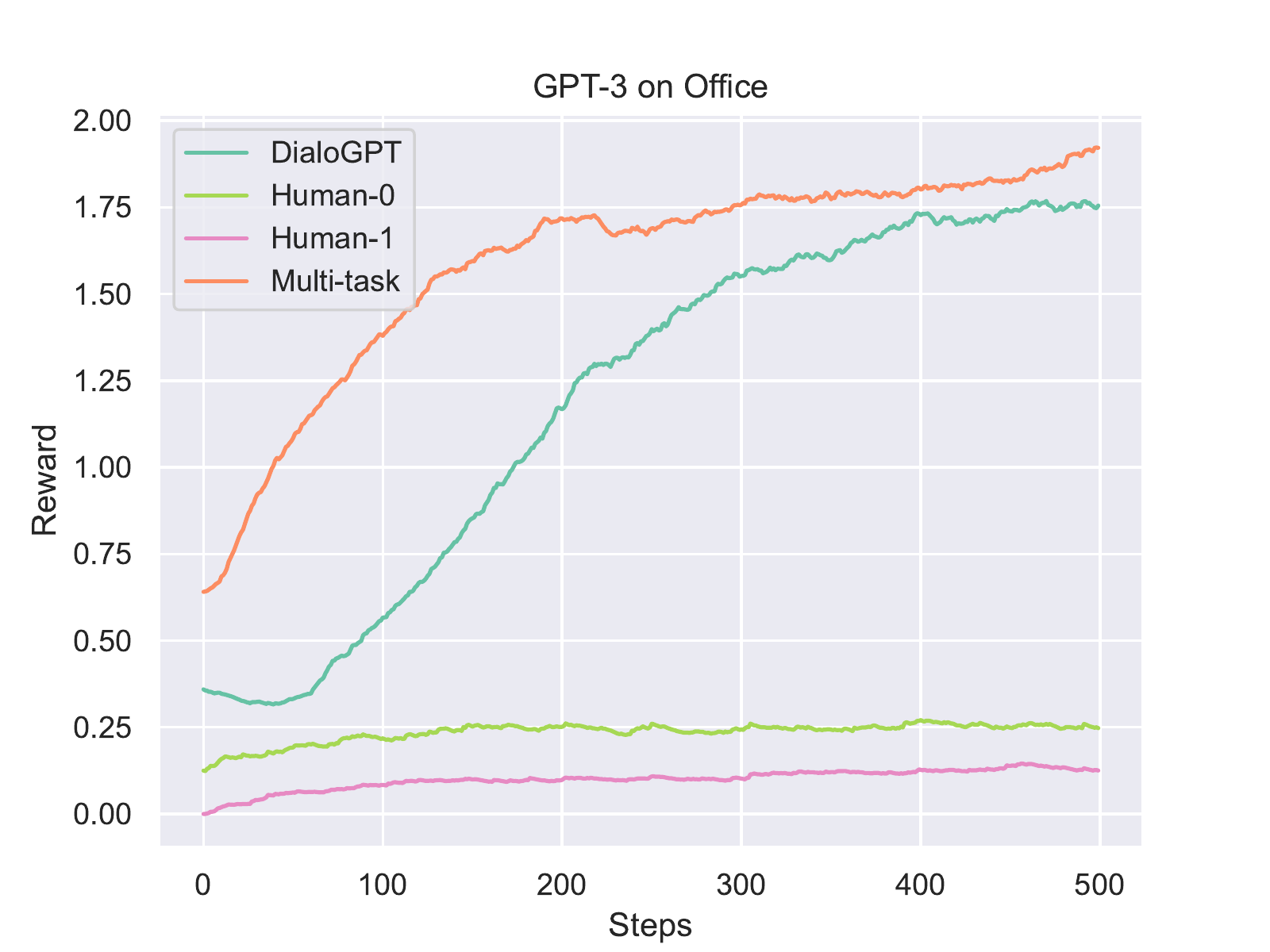}
        \caption{GPT-3 on Office}
        \label{fig:subim42}
    \end{subfigure}
    \caption{The result of GPT-3 evaluation.}
    \label{fig:image5}
\end{figure*}

\subsection{Emotion v.s. Topic}
The reward of the proposed model is higher that of the baseline in both controllable factors. Yet we can see that the coherence score decreases as the reward increases. For the model, the coherence score is an acceptable trade-off for the reward. Moreover, the coherence scores of the emotion are much higher than the ones of the topic. It is because the topic controllable factor is highly correlated to the context of the sentence. In order to raise the reward, the model is forced to say something about the given topic but irrelevant to the original context. In contrast, the emotion is relatively disentangled with the context, so it is possible to raise the reward while maintaining the coherence score.

\subsection{GPT-3 Evaluation}
To solid our proposed method on off-the-shelf conversational models, we also conduct the experiments on the OpenAI GPT-3 model. Figure~\ref{fig:image5} shows the promising result. It turns out that our proposed RL method can successfully steer the GPT-3 model API. Furthermore, The model with multi-task pre-training also outperforms the baseline model under the few-shot scenario.  

The same phenomenon can also be observed in table \ref{tab:table2}, the model with multi-task pre-training indeed achieves higher reward compared to the baseline. Furthermore, our proposed method also maintains the comparable coherence, perplexity, and diversity.

\begin{table*}[htbp]\small
        \centering
        \begin{tabular}{ccccccccccc}
        \hline
         &  &  &  & \multicolumn{2}{c}{\textbf{Pride}} &  &  & \multicolumn{2}{c}{\textbf{Office}} &  \\ \hline
         & \textbf{Method} & \textbf{Model} & \textbf{Reward} & \textbf{PPL} & \textbf{Coh.} & \textbf{SB-2} & \textbf{Reward} & \textbf{PPL} & \textbf{Coh.} & \textbf{SB-2}\\ \hline
         \multirow{9}{*}{\textbf{GPT-3}}  & \multirow{2}{*}{0 shot} & DGPT & 0.007 & 8.425 & 0.722 & 0.883 & 0.055 & 8.533 & 0.72 & 0.883 \\
         &  & Multi & \textbf{0.008} & 8.245 & 0.727 & 0.89 & \textbf{0.218} & 9.884 & 0.719 & 0.874 \\
         & \multirow{2}{*}{10 shot} & DGPT & \textbf{0.008} & 8.492 & 0.724 & 0.888 & 0.06 & 8.886 & 0.695 & 0.873\\
         &  & Multi & \textbf{0.008} & 8.173 & 0.72 & 0.888 & \textbf{0.374} & 13.385 & 0.761 & 0.861\\
         & \multirow{2}{*}{500 shot} & DGPT & 0.714 & 17.232 & 0.861 & 0.859 & 0.471 & 14.005 & 0.465 & 0.933\\
         &  & Multi & \textbf{0.789} & 3.748 & 0.416 & 0.947 & \textbf{0.544} & 19.474 & 0.84 & 0.858\\ \cline{2-11}
         & Human-0 & DGPT & 0.023 & 12.304 & 0.788 & 0.834 & 0.072 & 15.187 & 0.812 & 0.823 \\
         & Human-1 & DGPT & 0.067 & 14.916 & 0.714 & 0.86 & 0.041 & 13.703 & 0.676 & 0.871 \\ \cline{2-11}
         & no prompt & - & 0.0371 & 9.793 & 0.755 & 0.888 & 0.1 & 9.793 & 0.755 & 0.888  \\ \hline
        \end{tabular}

    \caption{\label{tab:table2}The result of few-shot testing on GPT-3. The column \textbf{Model} means which model the prompt generator was initialized from.}
\end{table*}

\begin{figure*}[h]
    \begin{subfigure}{0.5\linewidth}
        \includegraphics[width=1\linewidth]{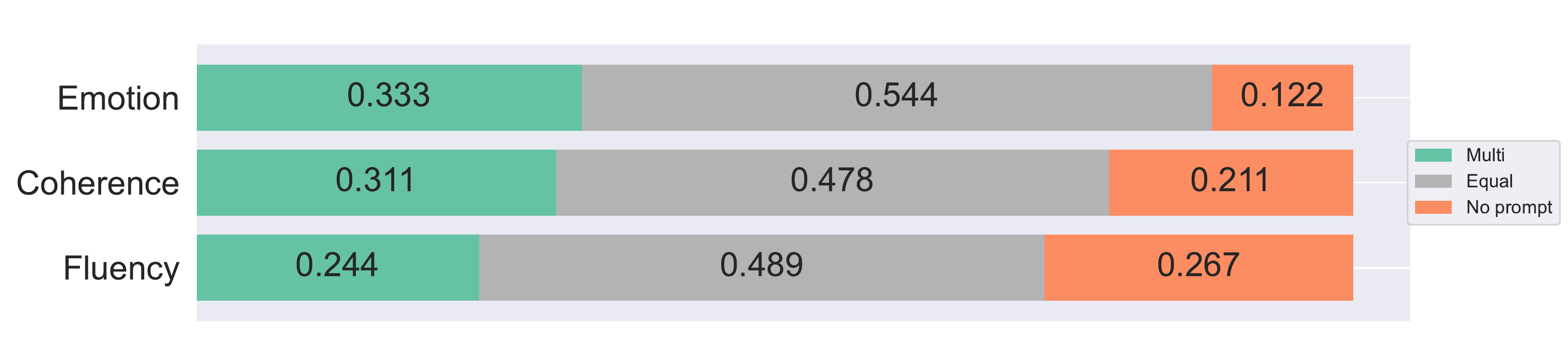} 
        \caption{Blenderbot on Emotion}
        \label{fig:subim41}
    \end{subfigure}
    \begin{subfigure}{0.5\linewidth}
        \includegraphics[width=1\linewidth]{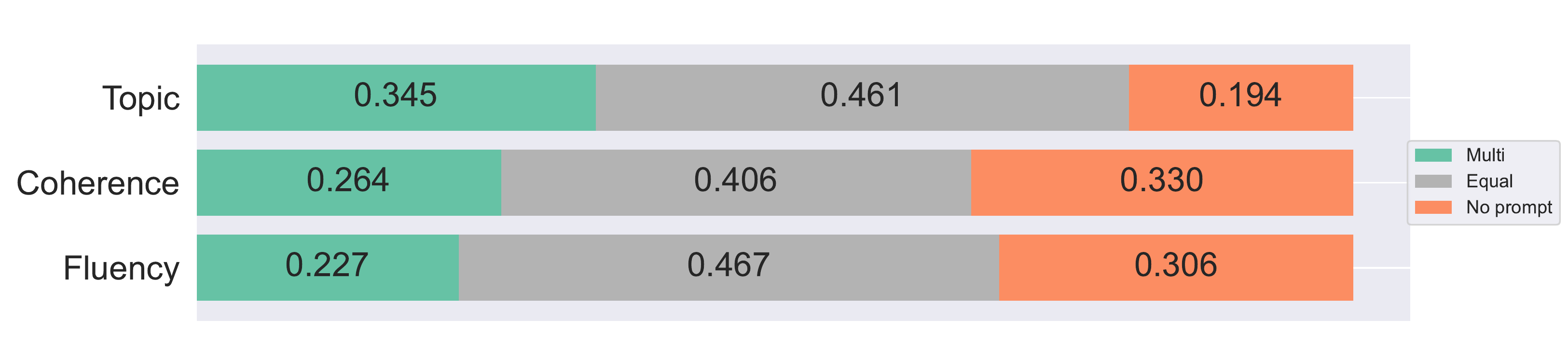}
        \caption{Blenderbot on Topic}
        \label{fig:subim42}
    \end{subfigure}
    \begin{subfigure}{0.5\linewidth}
        \includegraphics[width=1\linewidth]{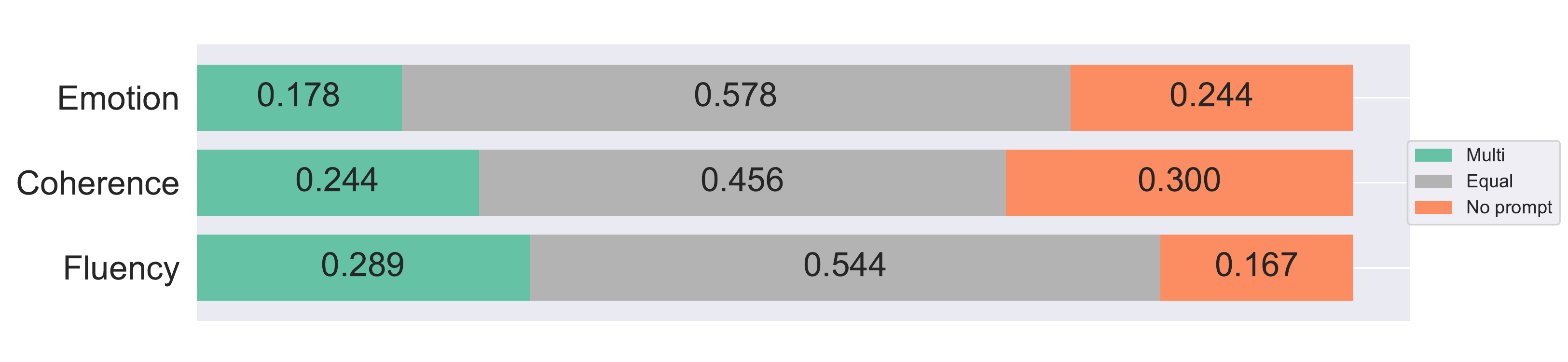} 
        \caption{DialoGPT on Emotion}
        \label{fig:subim43}
    \end{subfigure}
    \begin{subfigure}{0.5\linewidth}
        \includegraphics[width=1\linewidth]{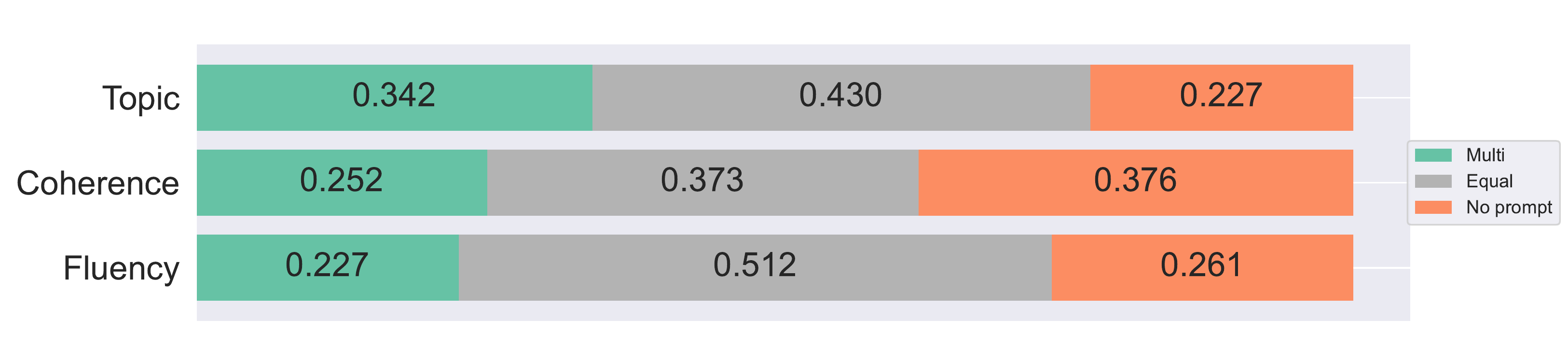}
        \caption{DialoGPT on Topic}
        \label{fig:subim44}
    \end{subfigure}
    \begin{subfigure}{0.5\linewidth}
        \includegraphics[width=1\linewidth]{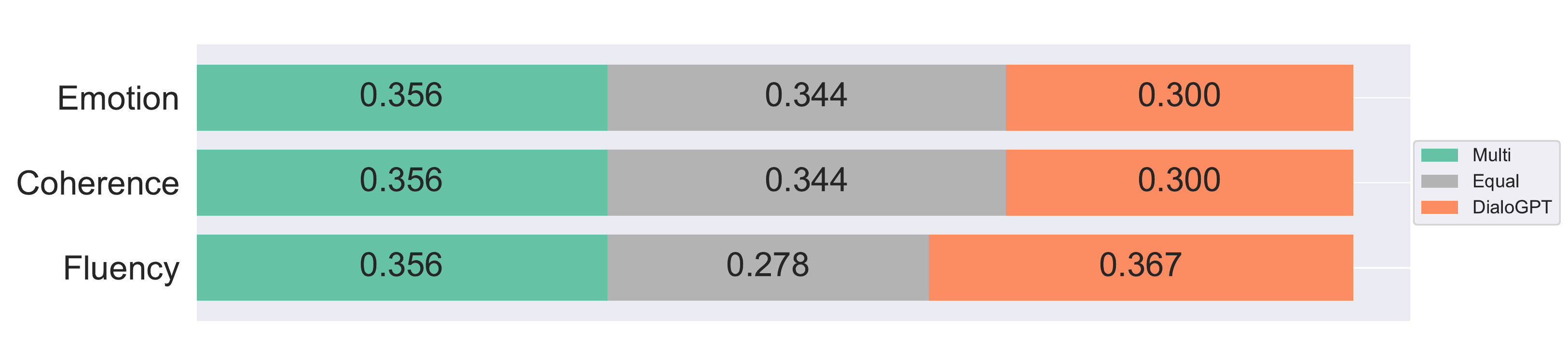} 
        \caption{Blenderbot on Emotion}
        \label{fig:subim31}
    \end{subfigure}
    \begin{subfigure}{0.5\linewidth}
        \includegraphics[width=1\linewidth]{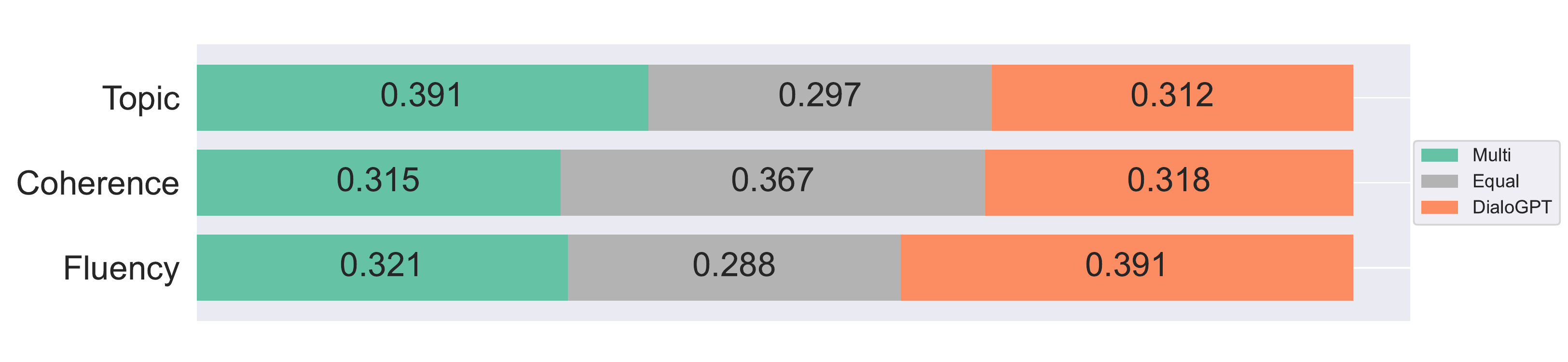}
        \caption{Blenderbot on Topic}
        \label{fig:subim32}
    \end{subfigure}
    \begin{subfigure}{0.5\linewidth}
        \includegraphics[width=1\linewidth]{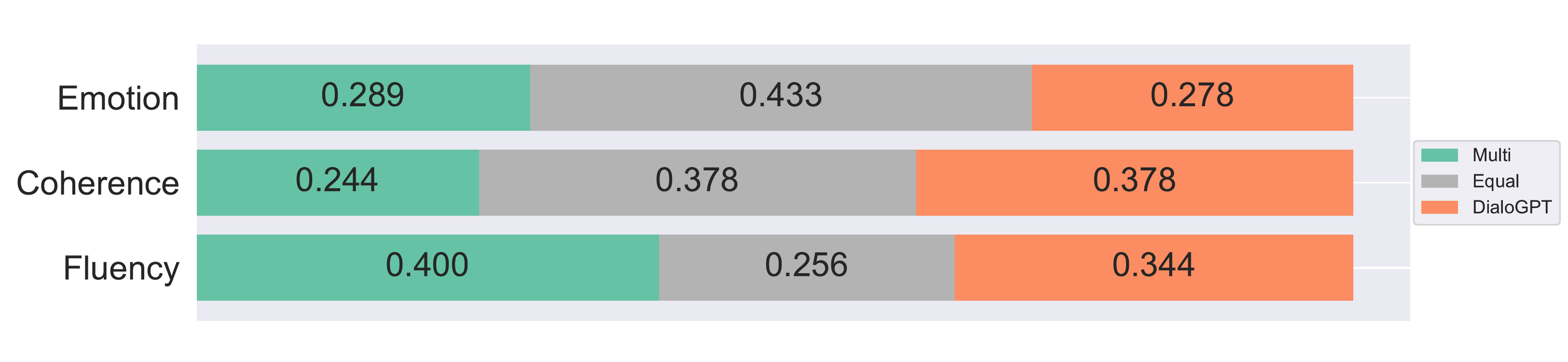} 
        \caption{DialoGPT on Emotion}
        \label{fig:subim33}
    \end{subfigure}
    \begin{subfigure}{0.5\linewidth}
        \includegraphics[width=1\linewidth]{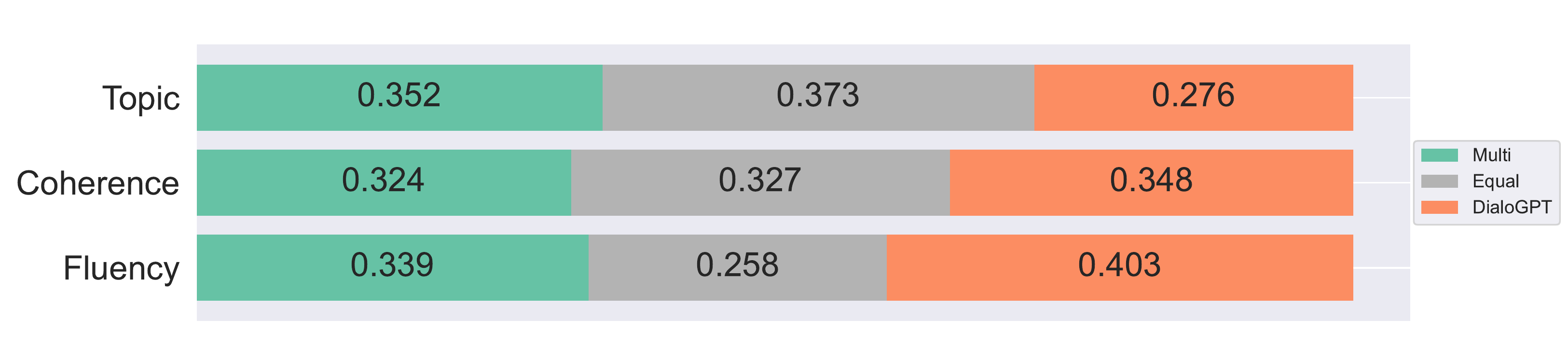}
        \caption{DialoGPT on Topic}
        \label{fig:subim34}
    \end{subfigure}
    \caption{The result of human evaluation. The number in the stack plot stands for the percentage that the corresponding model is rated higher. (a)-(d) are the results of Multi vs No prompt, (e)-(h) are the results of Multi vs DIaloGPT.}
    \label{fig:image4}
\end{figure*}

\subsection{Human Evaluation}
In our human evaluation, we conduct the experiments in two aspects. One is to compare if the multi-task learning method can successfully reduce the training steps compared to the model trained without multi-task learning. Another one is that we compare if our model can successfully steer models' response.
The results are shown in figure \ref{fig:image4}. In terms of the reward, out method indeed outperforms the model without any prompt input and the model without multi-task learning. Furthermore, we can also maintain the performance in the quality, coherence and diversity.

\section{Conclusion}
In this paper, we first proposed a prompt generator that trained with RL to steer off-the-shelf model's generation. Second, we further proposed a training technique that employed the concept of multi-task learning to speed up the convergence speed and enhance generator's generalizability on the unseen tasks.
In our proposed method, we designed two controllable factors for the prompt generator to generate prompts that can steer chatbots. One is that we aimed to control chatbot's response in various emotion reactions. Another one is that we let chatbot reply with responses that include the words under specific topics.
The experiment results demonstrate that the proposed method can successfully steer the SOTA models included DialoGPT, BlenderBot and GPT-3 without accessing their parameters. Besides, the convergence speed of model with multi-task learning is also faster than the one that did not train with multi-task learning.

\section*{Ethics}
In this paper, we proposed a learning framework that trains a prompts generator to influence off-the-shelf chatbots. We defined several rewards to reflect different behaviors that the generator can steer the chatbots.

We undertook this work because we envisioned a future in which a chatbot can be widely used and become  a digital companion for humans.
To that end, we need to put more efforts on extending the potential of present off-the-shelf chatbot.
Since our framework is reward-agnostic that could be optimize for any reward, we also expect that the experts could customize the profession reward definitions in their fields to bring the technique to higher level usage.

However, we also acknowledge the potential that this technique could be misused. Using our framework, ill-intentioned people could train chatbots with negative intentions and could threaten the stability of our society. 
% In the following sections, we first discussed possible scenarios of misusing our proposed methodology, and then we proposed ways to tackle them.
For example, we have identified the following means by which a malicious actor could take advantage of our proposed technology:
%We discussed the ethical considerations as below.

\begin{itemize}
\item \textbf{Emotional Manipulation}: One could train a prompt generator that control the chatbots with arousing negative emotions such as anxiety, sadness, or anger to wreak havoc on the human's mental state.
\item \textbf{Social Antagonism}: One could train a prompt generator with the "Topic Reward" to steer the chatbots to exhibit gender biases or use racist terms to purposefully destabilize society. 
\item \textbf{Political Interference}: One could train a prompt generator that induce the chatbots with the malicious intentions of manipulating the public’s political opinion.

\end{itemize}
\hfill \break
To prevent the aforementioned abuse of our method, we propose the following methods to counter them.
\begin{itemize}

    \item \textbf{Intention Classifier}: We could train a dialogue classifier that classifies whether a chatbot is purposefully controlled for malicious factors. We believe this is technically achievable as we could find many works that aim to distinguish whether a sentence is generated by humans or not \cite{gao2020dialogrpt}. To further refine this work, we could easily collect training datasets for this classifier by interacting with chatbots trained by our framework and other general-purpose chatbots. By doing this, we could inform humans when we detect that the chatbot they are conversing with is being manipulative.
\item \textbf{Special Token}: In the future, biomimetic technologies could blur the boundary between a living being and an artifact. We suggest that if the chatbot model generates the sentences, the sentence needs to be labeled with some special flag to tell people whether the chatbot generates the sentence with the intention. For instance, we can add  “<chatbot | intention>” before any chatbot’s response with the intention to inform people that a chatbot is trying to influence them. This will make users aware that they are interacting with a chatbot and can undermine the effectiveness of a malevolent attack.  
    \item \textbf{Safety Layer}: Inspired by \cite{adiwardana2020humanlike}, we could use a safety layer (e.g., an additional classifier) to filter out sensitive or toxic responses from chatbots during inference.
\end{itemize}

\paragraph{Future Work} To avoid malicious actors taking our framework and train their own prompt generator. The development of the \textbf{Intention Classifier} become an essential research topic. In future work, we would set the development of the Intention Classifier as the top priority. 
The functions of the  Intention Classifier are not only detect the intention of a dialogue system, it can also have an ability to generalize to any other dialogue systems. 
With the power of Meta-Learning~\cite{pmlr-v70-finn17a} the classifier is expected to train on a dialogue system with few data and could have the ability to detect whether sentences generated by the dialogue system are with intention. 

\paragraph{}

As developers of emerging technologies, we also take responsibility for defining the boundaries of these technologies. We will continue to refine the aforementioned methods to ensure that the proposed methodology improves public welfare as we intend it to.

\bibliography{anthology,custom}
\bibliographystyle{acl_natbib}

\appendix

\section{Appendix}
\label{sec:appendix}
\subsection{Computing Infrastructure}
\paragraph{CPU} 
Intel(R) Xeon(R) Gold 6154 CPU @ 3.00GHz * 4
\paragraph{Memory}  790964940 kB
\paragraph{GPU} Tesla V100-SXM2 32 GB
\paragraph{Numbers of Parameters} There are around 1.5 billion parameters in GPT-2 model.
\subsection{Detail of the tasks}
\paragraph{Task name} Table~\ref{table:5},~\ref{table:6},~\ref{table:3},~\ref{table:4} shows each emotion and topic task name. Also, we split all task into train tasks and test tasks, which they don't have any overlapping.

\begin{table}
\begin{tabular}{|c|c|}
\hline
Type & Topic  \\
\hline
\begin{tabular}[x]{@{}c@{}}Train\\Tasks\end{tabular} & \begin{tabular}[x]{@{}c@{}}vacation, postal, fish, amphibian, money, \\
car, thanksgiving, summer, military,\\
constitution, country, buildings,\\
fire, dogs, boat, flagday,\\
colors, virtues, columbus,\\
householddevices, baseball, reptiles \\
ocean, food, herbs, tools, irregularverbs,\\
fruit, christmas, nounandverb,\\
castle, stpatrick, art, fall, \\
halloween, cooking, weather, sewing, \\
groundhogday, body, presidentsday, \\
winter, newyear, mothersday, leaders, \\
dentist, driving, house, insect, time, \\
geography, people, jobs, carnival, energy, \\
chinesenewyear, camping, tree, spring, \\
container, sports, math, metals, \\
carparts, many, mammal, beach, \\
circus, pirate, desserts, biomes, birthday, \\
hats, rooms, furniture, sciences, clothes, \\
big, plants, computer, valentine, stores \\
landforms, bathroom, cookingtools, \\
goodluck, doctor, restaurant, \\
happiness, cats, bodiesofwater, \\
yard, astronomy, legal, kitchen,\\
\end{tabular} \\
\hline
Count & 95 \\
\hline
\end{tabular}
\caption{The number of train tasks for topic type, and each train task name.}
\label{table:3}
\end{table}

\begin{table}
    \centering
    \begin{tabular}{|c|c|}
    \hline
    Type & Topic  \\
    \hline
    \begin{tabular}[x]{@{}c@{}}Test\\Tasks\end{tabular} & \begin{tabular}[x]{@{}c@{}}election, longe, musicalinstruments,\\
    office, animal, water, science, shapes, \\
    school, transportation, dance, \\
    mythicalbeasts, family, vegetables, \\
    aprilfool, shoes, emotions, flowers, \\
    weapons, rocks, roadways,farm,\\
    \end{tabular} \\
    \hline
    Count & 22 \\
    \hline
    \end{tabular}
    \caption{The number of test tasks for topic type, and each test task name.}
    \label{table:4}
\end{table}

\begin{table}
    \centering
    \begin{tabular}{|c|c|}
    \hline
    Type & Emotion  \\
    \hline
    \begin{tabular}[x]{@{}c@{}}Train\\Tasks\end{tabular} & \begin{tabular}[x]{@{}c@{}}admiration, amusement, disapproval, \\
    disgust, embarrassment, excitement,love, \\
    fear, gratitude, grief, nervousness,\\
    anger, optimism,\\
    realization, relief,\\ 
    remorse, surprise, caring, \\
    curiosity, desire, disappointment\\
    \end{tabular} \\
    \hline
    Count & 21 \\
    \hline
    \end{tabular}
    \caption{The number of train tasks for emotion type, and each train task name.}
\label{table:5}
\end{table}

\begin{table}
    \centering
    \begin{tabular}{|c|c|}
    \hline
    Type & Emotion  \\
    \hline
    \begin{tabular}[x]{@{}c@{}}Test\\Tasks\end{tabular} & \begin{tabular}[x]{@{}c@{}}annoyance, pride, joy, sadness, approval,\\
    confusion,\\
    \end{tabular} \\
    \hline
    Count & 6 \\
    \hline
    \end{tabular}
    \caption{The number of test tasks for emotion type, and each test task name.}
\label{table:6}
\end{table}

\end{document}